\pdfoutput=1

\documentclass[11pt]{article}

\usepackage{emnlp2021}
\usepackage{times}
\usepackage{amssymb}
\usepackage{latexsym}

\usepackage{microtype}
\usepackage{tikz}
\usepackage{tikz-3dplot}
\usepackage{amsfonts}
\usepackage{multirow}
\usepackage{graphicx}
\usepackage{soul}
\usepackage{circuitikz}
\usepackage{dsfont}
\usepackage{commath}
\usepackage{amsmath}
\usepackage{mathtools}
\usepackage[mathscr]{eucal}
\usepackage{latexsym, amsmath, amssymb}

\usepackage[utf8]{inputenc}
\usepackage{pgfplots}
\DeclareUnicodeCharacter{2212}{−}
\usepgfplotslibrary{groupplots,dateplot}
\usetikzlibrary{patterns,shapes.arrows}
\pgfplotsset{compat=newest}
\usepackage{placeins}
\usepackage{subcaption}
\usepackage{mwe}
\usepackage{pifont}
\usepackage{xcolor}
\usepackage{soul}
\usepackage{paralist}
\usepackage{enumitem}
\usepackage{bbm}
\usepackage{cleveref}

\usepackage{booktabs}
\usepackage{relsize}
\usepackage{array}

\usepackage{hhline}
\usepackage{pgf}
\usepackage{multirow}
\usepackage{blkarray, bigstrut}
\usepackage{comment}

\newcommand{\method}[2][]{\textup{#2}\ensuremath{_{\mathtt{#1}}}\xspace}
\newcommand{\indep}{\method[INDEP]{G}}
\newcommand{\inter}{\method[INTER]{G}}
\newcommand{\gerry}{\method[GERRY]{G}}
\newcommand{\inlpindep}{\method[INDEP]{INLP}}
\newcommand{\inlpinter}{\method[INTER]{INLP}}
\newcommand{\inlpgerry}{\method[GERRY]{INLP}}

\newcommand{\conindep}{\method[INDEP]{CON}}
\newcommand{\coninter}{\method[INTER]{CON}}
\newcommand{\congerry}{\method[GERRY]{CON}}

\newcolumntype{E}{>{\collectcell}m{0.4cm}<{\endcollectcell}}  

\newcommand{\group}[2][]{\ensuremath{#1\mathcal{#2}}\xspace}
\newcommand{\white}{\group{W}}
\newcommand{\notwhite}{\group[\neg]{W}}
\newcommand{\male}{\group{M}}
\newcommand{\female}{\group{F}}

\Crefname{equation}{Eq}{Eqs}

\title{Evaluating Debiasing Techniques for Intersectional Biases}
\author{Shivashankar Subramanian \quad Xudong Han \\ {\bf Timothy
  Baldwin \quad Trevor Cohn \quad Lea Frermann} \\School of Computing and Information Systems\\ University
  of Melbourne, Australia \\
 {\small \url{shivashankarrs@gmail.com} \quad \url{xudongh1@student.unimelb.edu.au}} \\ {\small \url{{tbaldwin, t.cohn, lfrermann}@unimelb.edu.au}}}

\date{}

\begin{document}
\maketitle

\begin{abstract}
Bias is pervasive in NLP models, motivating the development of
automatic debiasing techniques. Evaluation of NLP debiasing methods has
largely been limited to binary attributes in isolation, e.g., debiasing
with respect to binary gender or race, however many corpora involve
multiple such attributes, possibly with higher cardinality.
In this paper we argue that a truly fair model must consider `gerrymandering' groups which comprise not only single attributes, but also intersectional groups.
We evaluate a form of bias-constrained model which is new to NLP, as
well an extension of the iterative nullspace projection technique which
can handle multiple protected attributes.
\end{abstract}

\section{Introduction}

Text data reflects the social and cultural biases in the world, and NLP models and applications trained on such data have been shown to reproduce and amplify those biases. Discrimination has been identified across diverse sensitive attributes including gender, disability, race, and religion \cite{caliskan2017semantics, may2019measuring, garimella2019women, nangia2020crows, li2020unqovering}. While early work focused on debiasing typically binarized protected attributes in isolation (e.g., age, gender, or race;~\newcite{caliskan2017semantics}), more recent work has adopted a more realistic scenario with multiple sensitive attributes~\cite{li2018towards} or attributes covering several classes \cite{manzini2019black}. 




In the context of multiple protected attributes, {\it gerrymandering} refers to the phenomenon where an attempt to make a model fairer towards some group results in increased unfairness towards another group~\cite{buolamwini2018gender,kearns2018preventing,yang2020fairness}.
Notably, algorithms can be fair towards independent groups, but not towards all intersectional groups.
Despite this, debiasing approaches within NLP have so far been evaluated only using \textit{independent group fairness} when modelling datasets with multiple attributes, disregarding intersectional subgroups defined by combinations of sensitive attributes (see \Cref{figure:priv}).

The primary goal of this work is to evaluate independent and intersectional identity debiasing approaches in relation to fairness gerrymandering for text classification tasks. To this end, we evaluate bias-constrained models \cite{cotter2019two} and iterative nullspace projection (\textsc{INLP}; \citet{ravfogel2020null}), a post-hoc debiasing method which we extend to handle intersectional groups. The constrained model jointly optimizes model performance and model fairness, while INLP seeks to learn a hidden representation which is independent of the protected attributes. INLP does not consider the trade-off between accuracy and fairness, but rather it iteratively maximizes fairness 
in an unconstrained fashion. 



In this work, we address the following questions:
\begin{compactitem}
    \item Are debiasing approaches based on independent groups more prone to fairness gerrymandering than methods using intersectional groups?
    \item How do INLP and bias-constrained approaches,
    and their extensions to handle intersectional groups, fare compare in terms of both predictive accuracy and fairness?
\end{compactitem}

\section{Background}
\label{sec:background}


\begin{figure}
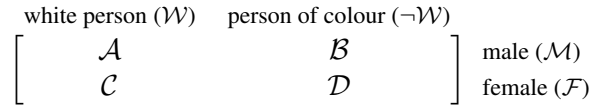

\begin{equation*}
  \begin{blockarray}{*{2}{c} l}
    \begin{block}{*{2}{>{$\footnotesize}c<{$}} l}
      white person (\white) & person of colour (\notwhite)\\
    \end{block}
    \begin{block}{[*{2}{c}]>{$\footnotesize}l<{$}}
      \mathcal{A} & \mathcal{B} & male (\male)\\
      \mathcal{C} & \mathcal{D}  & female (\female)\\
    \end{block}
  \end{blockarray}
\end{equation*}
\vspace{-3ex}
\caption{Group intersection and gerrymandering: $\mathcal{A}$=white male, $\mathcal{B}$=male person of colour, $\mathcal{C}$=white female, $\mathcal{D}$=female person of colour.}
\label{figure:priv}
\end{figure}

Debiasing with respect to more than a single protected attribute ($|Z|>1$) requires grouping data points according to their associated protected values. In this work, debiasing approaches are trained \emph{wrt} the three settings detailed below ~\cite{kearns2018preventing,yang2020fairness}, evaluated based on aggregated violations over \gerry, which is a commonly used evaluation strategy for intersectional identities \cite{yang2020fairness}.


\vspace{-0.2cm} 
\paragraph{\textbf{Independent groups (\indep)}} Each private attribute is treated as an independent group. In the case of all binary attributes, this gives 2$|Z|$ groups. As illustrated in Figure \ref{figure:priv}, for binary gender and race,\footnote{Higher cardinality attributes are also supported.} \female, \male, \white, and \notwhite are four independent groups, which are overlapping, e.g., with \female  referring to both white females and female persons of colour ($\mathcal{C}$ and $\mathcal{D}$). 
\vspace{-0.2cm} 
\paragraph{\textbf{Intersectional groups (\inter)}} Non-overlapping, exhaustive combinations of groups, which in the binary attribute case gives 2$^{|Z|}$ combinations. For our example in Figure~\ref{figure:priv}, the intersectional groups are $\mathcal{A}$, $\mathcal{B}$, $\mathcal{C}$ and $\mathcal{D}$.

 \vspace{-0.2cm} 
 \paragraph{\textbf{Gerrymandering intersectional groups (\gerry)}}
Overlapping subgroups are defined by group intersections of any subset of private attributes (3$^{|Z|}$ for binary attributes, where the 3rd value denotes a wildcard). For the example in Figure~\ref{figure:priv}, the \gerry groups are \female, \male, \white, \notwhite, $\mathcal{A}$, $\mathcal{B}$, $\mathcal{C}$, and $\mathcal{D}$. 


\section{Proposed Approaches}

We aim to learn a model to predict main task labels (e.g., sentiment) from input data $X$ (e.g., reviews) by means of a learnt hidden 
representation $X_h$. $X_h$ 
will generally encode implicit biases, and our goal is to predict the output without the influence of social identities encoded in~$X_h$. We assume  access to the protected attributes for each instance.
Below, we detail the two primary models targeted in this work: INLP, and the bias-constrained model.

\subsection{INLP}
\label{sec:inlp}

INLP is applied to frozen hidden representations, where we first learn a linear classifier $\mathbf{W}$ using $X_h$ as the independent variables to predict a protected attribute (binary- or multi-valued). Next, $X_h$ is projected onto the nullspace of $\mathbf{W}$ (denoted $P_{N(\mathbf{W})}X_h$) to remove the protected information, which forms the input for learning a classifier of the main task label. In real-world scenarios, bias may arise due to a combination of multiple protected attributes, hence we extend the single attribute setting of \citet{ravfogel2020null} to this situation. 
Assuming $k$~protected attributes, and classifiers trained to predict those denoted as $\{ \mathbf{W}^i \}_{i=1}^k$, we compute the intersection of nullspaces \cite{strang2014linear} across identities,
\begin{equation} 
\bigcap_{i=1}^k N(\textbf{W}^i) =
N\left(\left[\begin{matrix}\mathbf{W}^1 \\ \mathbf{W}^2 \\ \vdots \\ \mathbf{W}^k\end{matrix}\right]\right) \, .
\label{eqn:strang}
\end{equation}
No special properties are imposed over $\mathbf{W}^i$ (e.g., orthogonality), which allows for parallel training. 


Even after debiasing by projecting $X_h$ 
onto the nullspace of classifiers, hidden representations can still retain protected information \cite{ravfogel2020null}. Hence we follow \citet{ravfogel2020null} and resort to iteratively removing bias, rather than doing it in a single step. Nullspace projection will reduce the rank of data rapidly with intersectional identities\footnote{The rank reduces by 1 for each protected attribute.} (see Figure \ref{fig:inlp-pvsnp}). To circumvent this issue, we project onto the nullspace of the \emph{principal component} of the bias subspace ($N_p$). This is equivalent to projecting onto the nullspace of some of the intersectional identities in each iteration; similar strategies have been employed for multi-class supervised principal component analysis \cite{piironen2018iterative}. 


Following~\citet{ravfogel2020null}, while incorporating our principal component-based nullspace projection as described above, the intersection of nullspaces across iterations is computed as
%
\begin{equation} 
\bigcap_{i=0}^m N_p(\textbf{W}_i) 
= N\left(\sum_{i=0}^{m} P_{R'(\textbf{W}_i)} \right),
\label{eqn:benisrael}
\end{equation}
where $N_p$ denotes the nullspace based on the principal component of the bias subspace, $\textbf{W}_i$ is the set of classifiers trained in each iteration to predict the protected attributes, and $R'$ is the principal component of the rowspace computed by stacking classifier weight vectors across multiple protected attributes. Unlike \Cref{eqn:strang}, the formulation in \Cref{eqn:benisrael} requires orthogonality of subspaces across each iteration.


\subsection{Bias-constrained Model}
\label{sec:tfco}

A bias-constrained model combines the main task objective and fairness constraints, 
\begin{equation*}
\begin{aligned}
\min_{\theta} \quad & \ell (\mathcal{F} (x, \theta), y) \\
\textrm{s.t.} \quad & \forall g \in [G] \,\, \gamma_g \lvert \phi_g - \phi \rvert \leq \nu, \\
\end{aligned}
\end{equation*}
%
where $\ell (\mathcal{F} (x, \theta), y)$ denotes the primary objective (e.g., error rate);
$G$ are the groups defined in \Cref{sec:background}; $\phi_g$ denotes the group-wise performance; and $\phi$ denotes overall performance of the model (e.g., TPR). $\nu$ is a slack variable, which controls the maximum deviation allowed. 
$\gamma_g$ is the inverse proportion of positive examples in each group, $\frac{|g|}{|g_{y=1}|}$, meaning that the more underrepresented a group is \emph{wrt} some target class (say {\it toxicity}), the smaller its accepted deviation from the overall performance.

Each group-wise constraint is denoted as $\psi_g$. The constraints involve a linear combination of indicator variables, which is not differentiable \emph{wrt} $\theta$. 
A common approach to handle this constrained optimization problems is using the Lagrangian,
\begin{equation}
\begin{aligned}
 \mathcal{L}(\theta, \lambda) =  \ell (\mathcal{F} (x, \theta), y) + \sum_{g=1}^{|G|}  \lambda_g \psi_g \, ,\\
\end{aligned}
\label{eqn:tfco}
\end{equation}
which is minimized over $\theta$ and maximized over $\lambda$. Similar formulations have been used for learning fair models with structured data \cite{cotter2019two, yang2020fairness, 10.5555/3322706.3362016}. In this work, we apply this method to NLP tasks and use the two-player zero-sum game approach for optimization, where the first player chooses $\theta$ to minimize $\mathcal{L}(\theta, \lambda)$, and the second player enforces fairness constraints by maximizing $\lambda$ \cite{kearns2018preventing, cotter2019two, yang2020fairness}. Specifically we use the implementations available in TensorFlow constrained optimization \cite{cotter2019training, cotter2019two}.\footnote{
\url{https://github.com/google-research/tensorflow_constrained_optimization}}

The approach iterates over $\mathtt{T}$ iterations, where the primal player updates $\theta$ with a fixed $\lambda$, and dual player updates $\lambda$ with a fixed $\theta$. This results in multiple models, $\{\theta_1, \theta_2, ... \theta_{\mathtt{T}}\}$ and $\{\lambda_1, \lambda_2, ... \lambda_{\mathtt{T}}\}$, which trade off performance and fairness. We use the Adam optimizer \cite{kingma2014adam} with learning rate $10^{-3}$.


\section{Experimental Results}
\label{sec:faireval}

We evaluate using two binary classification datasets: (1) Twitter hate speech detection; and (2) occupation classification using biography text. For INLP we use logistic regression to construct $\mathbf{W}$. All the approaches have hyperparameters controlling the performance--fairness trade-off, and as no choice of hyperparameter will optimize both objectives simultaneously, we adopt the concept of \textit{Pareto optimality} \cite{godfrey2007algorithms}.

\subsection{Settings}
\label{sec:settings}

For INLP we consider different definitions of the bias subspace,
which we illustrate \emph{wrt} Figure \ref{figure:priv}: 
(a) two linear classifiers, trained to discriminate between independent groups \indep:
\female and \male, \white and \notwhite; 
(b) four linear classifiers to discriminate between intersectional groups \inter: $\mathcal{A}$,
$\mathcal{B}$, $\mathcal{C}$, and $\mathcal{D}$ (one vs.\ rest); 
and (c) eight linear classifiers to discriminate between \gerry groups. 
We refer to the three approaches as \inlpindep, \inlpinter, and \inlpgerry respectively. For the constrained model,  constraints capturing the performance deviation of \indep, \inter, and \gerry are used in \Cref{eqn:tfco}, which we refer to as \conindep, \coninter, and \congerry, respectively.

\paragraph{Metrics} We assess a model's predictive performance using F-score, and fairness based on average-violations \cite{yang2020fairness} of true positive rate (\textit{equality of opportunity}), across all the \gerry groups: $\frac{1}{|\gerry|}\sum_{g = 1}^{\gerry} |\mathtt{tpr}_g - \mathtt{tpr}|$.


\subsection{Hate speech detection}
\label{exp:hs}

\begin{figure}[t]
   \centering
    \includegraphics[scale=0.5]{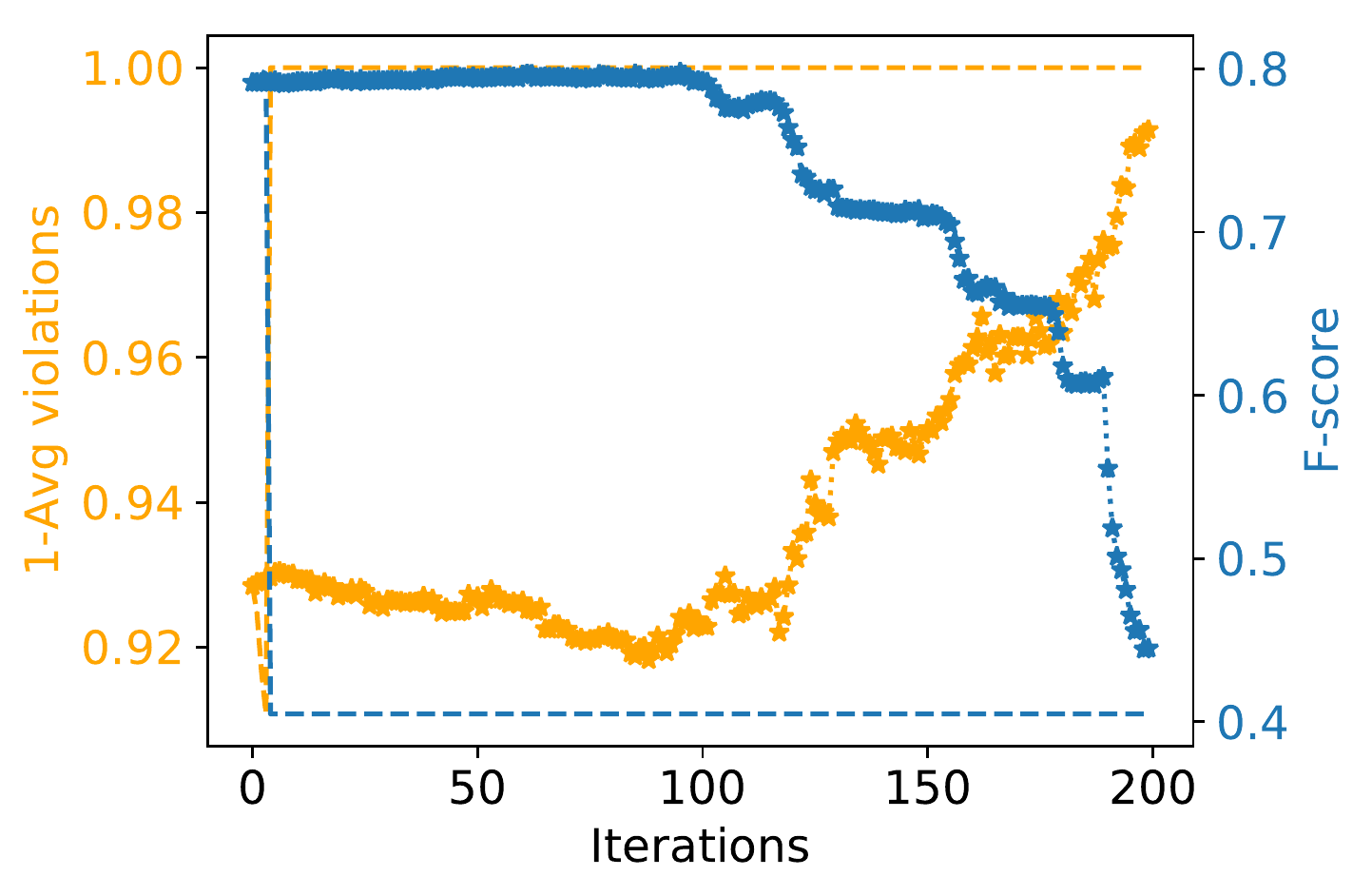}
\caption{F-score vs.\ fairness results for INLP (hatespeech dev set). Projecting onto the null space of the principal bias component (dotted line and stars) results in a better trade-off than directly projecting onto the nullspace of the entire bias sub-space (dashed line).}
   \label{fig:inlp-pvsnp}
\end{figure}

\begin{figure*}
\centering
\begin{subfigure}{.5\textwidth}
  \centering
  \includegraphics[scale=0.48]{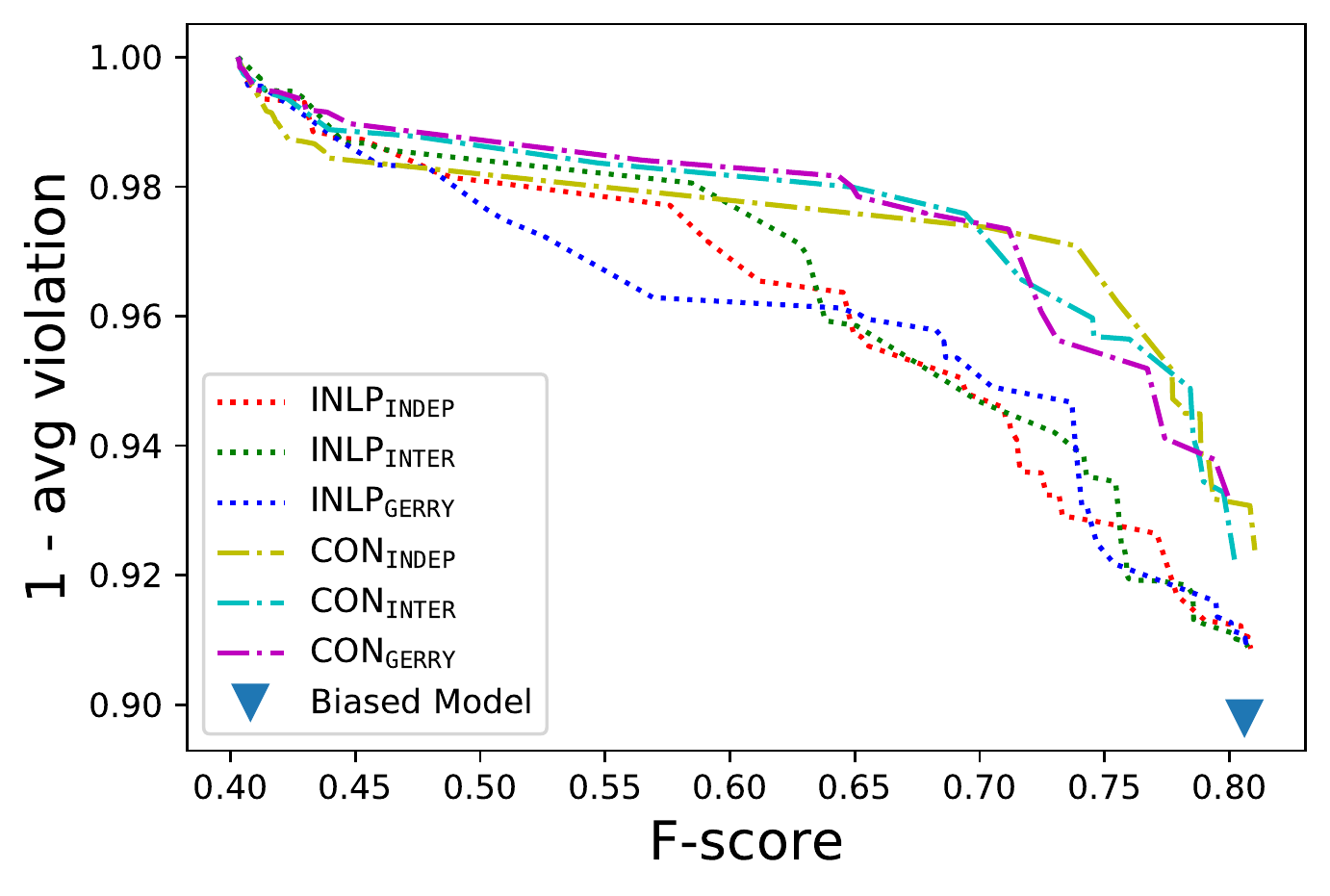}
  \caption{Hatespeech detection}
  \label{fig:hspareto}
\end{subfigure}%
\begin{subfigure}{.5\textwidth}
  \centering
  \includegraphics[scale=0.48]{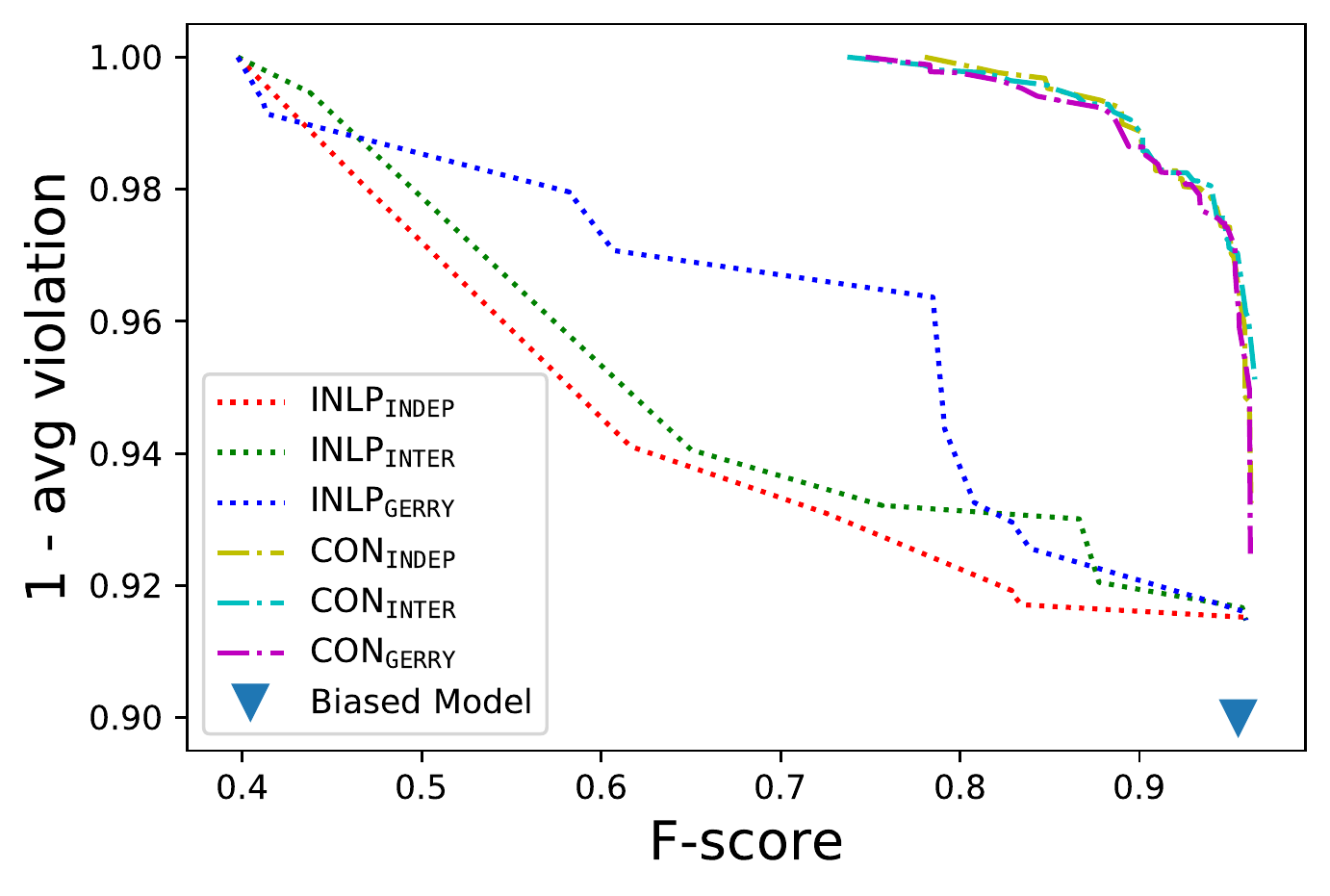}
  \caption{Occupation classification}
  \label{fig:biospareto}
\end{subfigure}
\caption{Pareto frontier model test results for the trade-off between $F_1$ score and fairness, measured as $1-$ the average TPR violation across all \gerry groups. The top-right represents ideal performance. }
\label{fig:pareto}
\end{figure*}


We use the English Twitter hate speech detection dataset of \citet{huang2020multilingual}, where each tweet is labeled with a binary hate speech label, as well as binary  demographic group identity indicators  for the tweet authors: gender (f or m), age ($\leq$ or $>$ median), location (US or not US), and race (white or other). We use the data splits from \citet{huang2020multilingual}, but further filter each partition to the subset of tweets for which all demographic author labels are available, resulting in
23K (train), 5K (dev), and 5K (test) tweets. 

We use a biGRU as our base model, following \citet{huang2020multilingual}. In \Cref{fig:inlp-pvsnp}, we show the benefits of our proposed INLP approach for debiasing gerrymandering intersectional identities, compared to the naive extension of INLP for multiple protected attributes (which projects onto the entire bias subspace). 
With 80 \gerry groups, the naive extension leads to a drastic drop in performance after 4 iterations (the dashed blue line), as in each iteration the rank is reduced by 80. On the other hand, the proposed approach achieves a much more nuanced trade-off between accuracy and fairness (the blue and yellow dotted line + stars, resp.).

We present the Pareto frontier for the different models on the test-set in \Cref{fig:hspareto}, by varying the hyper-parameters of the approaches\footnote{For INLP, \# iterations = 1 to dim($X_h$); for the constrained models, \# iterations ($\mathtt{T}$) = 1 to 100, $\nu$ = 1e-4 to 1, $\gamma_g$ = 1 or $\frac{|g|}{|g_{y=1}|}$.} each under of the three different debiasing settings, i.e.\ \indep, \inter, and \gerry. From the results, for F-score close to the base biGRU (= biased model), the independent model provides the best trade-off, while debiasing for intersectional identities provides a better trade-off overall, and constrained models perform better than INLP. 

\subsection{Occupation classification}
\label{exp:occ}


For our second experiment, we use the occupation classification dataset of \citet{de2019bias}. This comprises short web biographies, annotated by gender and profession. We augment the dataset with economic status (wealthy vs.~rest, using World Bank data) based on the country the individual is based in, labelled based on the first sentence of each bio, which we perform entity linking on using AIDA-light \cite{nguyen2014aida}, and map locations and institutions to countries based on their Wikipedia articles. We computed the accuracy of the economic status mapping by manually analyzing the output of over 200 biographies, resulting in a 93\% accuracy and 90\% agreement (Jaccard coefficient) between three annotators.   
We use a subset of two highly stereotyped classes of the dataset --- nurse vs.~surgeon --- resulting in 
13K/1.7K/5.4K instances in
train/dev/test.\footnote{90\% of surgeons are male and 87\% of nurses are female; 97.5\% of nurses are from wealthy economies.}

We use the BERT (base uncased) \cite{devlin2019bert} [CLS]-encoding followed by an MLP (300-d with ReLU activations) for classification. The Pareto results\footnote{Constrained models are run for $\mathtt{T}=1$ to 500 iterations.} on the test set
are shown 
in Figure \ref{fig:biospareto}. \inlpinter and \inlpgerry perform better than \inlpindep, and the constrained models once again perform best overall, with intersectional identities providing no significant gains over \conindep. Even in a highly stereotyped setting, constrained models achieve high predictive performance, in contrast to INLP which improves fairness but greatly reduces predictive performance, consistent with the single protected attribute results in \citet{ravfogel2020null}. 

\subsection{Constrained analysis}
\begin{table*}
    \centering
    \begin{subfigure}[c]{\textwidth}
    \small
    \centering
    \begin{tabular}{lcccccccc}
    \toprule
    && \multicolumn{3}{c}{\bf Trade-off 5\%} && \multicolumn{3}{c}{\bf Trade-off 10\%}\\
    \cmidrule(lr){3-5} \cmidrule(lr){7-9}
 \bf Approach && \bf F$_1$ & \bf Max violation & \bf Avg violation && \bf F$_1$ & \bf Max violation & \bf Avg violation  \\
    \midrule
  \inlpindep && 0.763 &0.333 &0.079 && 0.732 & 0.242 & 0.076 \\ 
   \inlpinter  && 0.785 & 0.312 & 0.083 &&   0.736 & 0.258 &0.064 \\ 
 \inlpgerry  &&  0.778 & 0.453 & 0.086 && 0.737 & 0.199 & 0.054  \\
\conindep && 0.777 & 0.230 & \textbf{0.048} && 0.724 & 0.274 & \textbf{0.031} \\
\coninter  &&  0.772 & 0.254 & 0.063 && 0.746 & 0.172 & 0.043 \\
\congerry  && 0.759 & \textbf{0.135} & 0.059 && 0.715 & \textbf{0.150} & 0.047 \\
\midrule
Biased model && 0.806 & 0.338 & 0.102&& 0.806 & 0.338 & 0.102\\
        \bottomrule
    \end{tabular}
    \caption{Hatespeech classification}
    \label{tab:hs}
    \end{subfigure}

    \vspace*{2mm}

    \begin{subfigure}[c]{\textwidth}
    \small
    \centering
    \begin{tabular}{lcccccccc}
    \toprule
        && \multicolumn{3}{c}{\bf Trade-off 5\%} && \multicolumn{3}{c}{\bf Trade-off 10\%}\\
    \cmidrule(lr){3-5} \cmidrule(lr){7-9}
 \bf Approach && \bf F$_1$ & \bf Max violation & \bf Avg violation && \bf F$_1$ & \bf Max violation & \bf Avg violation  \\
    \midrule
 \inlpindep && 0.949 & 0.291 & 0.087 &&  0.949 & 0.291 & 0.087  \\
\inlpinter  &&  0.956 &0.313 &0.090 && 0.956 &0.313 &0.090\\
\inlpgerry  &&  0.917 & 0.242 & 0.090 && 0.904 & 0.224 & 0.088\\
\conindep && 0.953 & \textbf{0.134} &\textbf{0.041} && 0.940 & \textbf{0.102} & \textbf{0.028}\\
\coninter  &&  0.949 & 0.225 & 0.065 && 0.908 & 0.110  & \textbf{0.028}\\
\congerry  &&  0.951 & 0.170 & 0.049 && 0.905 &0.137 &0.035\\
\midrule
Biased model && 0.955 & 0.350 & 0.100 && 0.955 & 0.350 & 0.100 \\
        \bottomrule
    \end{tabular}
        \caption{Occupation classification}
    \label{tab:biasbios}
    \end{subfigure}
\caption{Test performance of salient models on hatespeech and occupation classification. Least biased models within a given performance tradeoff thresholds are chosen from development set.}
\end{table*}

In addition to Pareto curves (Figure \ref{fig:pareto}), we compare the fairest models under a minimum performance constraint, where the most fair models (based on average violations) which exceed a performance threshold on the development set are chosen, and evaluated on the test set. This allows us to measure the generalizability of models occupying optimal Pareto region for the development set, to an unseen test set. We provide results under two constrained scenarios --- models with least bias on development set trading off 5\% and 10\% of predictive performances. In addition to average violation, we also report maximum violation as $\underset{g \in \text{\gerry}}{\max} |\mathtt{tpr}_g - \mathtt{tpr}|$, which measures the upper bound of unfairness towards any subgroup~\cite{yang2020fairness}.

For hatespeech classification, results are provided in Table \ref{tab:hs}. With a slack of up to 10\% tradeoff in predictive performance, \inlpgerry shows better predictive performance and less violations than both \inlpinter and \inlpindep. For the closer 5\% tradeoff, \inlpinter has higher F-score with a slightly worse avg violation and better max violation. The constrained models debiasing for \gerry are fairest (lowest max and avg violation). Similarly, for occupation classification (Table \ref{tab:biasbios}), \inlpgerry achieves lowest maximum violation among INLP models, while overall the constrained models, specifically \conindep, perform best. This confirms the conclusion based on Pareto plots that adding intersectional identities for INLP-based debiasing is beneficial, whereas for constrained models which directly regulate the expected performance \emph{wrt} different subgroups, they only provide mild gains.

\section{Conclusion and Future Work} 

We examined the impact of independent vs.\ intersectional groups on classifier fairness. We proposed an extended version of INLP, which we compared against bias-constrained models. INLP is a post-hoc debiasing method, while the constrained model requires full joint training. Debiasing in INLP happens in an unconstrained way, where sensitive information is removed from the latent representations independent of predictive performances, leading to fairer models at the cost of overall accuracy \cite{han2021diverse}. From a practical perspective, post-hoc debiasing may be the only option if it is impossible or undesirable to re-train a model from scratch with a regularized objective. Our paper shows that lighter-weight post-hoc debiasing is indeed possible in the case of overlapping groups, and quantifies the additional advantage of joint training and debiasing in the context of two data sets. The constrained models especially are more robust to stereotyped settings, and using intersectional identities benefits INLP in particular. 


\section*{Acknowledgements}
We thank Michael Twiton for his inputs on the INLP approach. This  work  was  funded  in  part by the Australian Government Research Training Program Scholarship, and the Australian Research Council.
\bibliographystyle{acl_natbib}
\bibliography{custom}

\end{document}